\documentclass[letterpaper]{article} 
\usepackage{aaai23}  
\usepackage{amsmath,amssymb} 
\usepackage{times}  
\usepackage{helvet}  
\usepackage{courier}  
\usepackage[hyphens]{url}  
\usepackage{graphicx} 
\usepackage{natbib}  
\usepackage{caption} 
\usepackage{algorithm}

\usepackage{graphicx}

\usepackage{tikz}
\usepackage{comment}
\usepackage{color}

\usepackage{xcolor}
\usepackage{adjustbox}
\usepackage{multirow}
\usepackage{algorithm}
\usepackage{algpseudocode}
\usepackage{booktabs}
\usepackage[title]{appendix}
\usepackage{newunicodechar}
\usepackage{wrapfig,lipsum,booktabs}

\newcommand{\scriptX}{\mathcal{X}}
\newcommand{\scriptY}{\mathcal{Y}}

\usepackage{xcolor,pifont}
\newcommand*\colourcheck[1]{%
  \expandafter\newcommand\csname #1check\endcsname{\textcolor{#1}{\ding{52}}}%
}
\colourcheck{blue}
\colourcheck{green}
\colourcheck{red}

\usepackage{newfloat}
\usepackage{listings}
\DeclareCaptionStyle{ruled}{labelfont=normalfont,labelsep=colon,strut=off} 
\floatstyle{ruled}
\newfloat{listing}{tb}{lst}{}
\floatname{listing}{Listing}
\title{Key Feature Replacement of In-Distribution Samples for Out-of-Distribution Detection}
\author {
    Jaeyoung Kim\equalcontrib\textsuperscript{\rm 1},
    Seo Taek Kong\equalcontrib\thanks{This work was submitted while the authors worked at VUNO Inc.}\textsuperscript{\rm 2},
    Dongbin Na\textsuperscript{\rm 1},
    Kyu-Hwan Jung\footnotemark[2]\textsuperscript{\rm 3}
}
\affiliations {
    \textsuperscript{\rm 1} VUNO, Inc. \\
    \textsuperscript{\rm 2} University of Illinois, Urbana-Champaign \\
    \textsuperscript{\rm 3} Samsung Advanced Institute for Health Sciences and Technology, Sungkyunkwan University \\
    jaeyoung.kim@vuno.co, skong10@illinois.edu, dongbin.na@vuno.co, khwanjung@skku.edu
}

\usepackage{bibentry}

\begin{document}


\maketitle


\begin{abstract}
    Out-of-distribution (OOD) detection can be used in deep learning-based applications to reject outlier samples from being unreliably classified by deep neural networks.
    Learning to classify between OOD and in-distribution samples is difficult because data comprising the former is extremely diverse.
    It has been observed that an auxiliary OOD dataset is most effective in training a ``rejection'' network when its samples are semantically similar to in-distribution images.
    We first deduce that OOD images are perceived by a deep neural network to be semantically similar to in-distribution samples when they share a common background, as deep networks are observed to incorrectly classify such images with high confidence.
    We then propose a simple yet effective \textbf{K}ey \textbf{I}n-distribution feature \textbf{R}eplacement \textbf{BY} inpainting (KIRBY) procedure that constructs a surrogate OOD dataset by replacing class-discriminative features of in-distribution samples with marginal background features.
    The procedure can be implemented using off-the-shelf vision algorithms, where each step within the algorithm is shown to make the surrogate data increasingly similar to in-distribution data.
    Design choices in each step are studied extensively, and an exhaustive comparison with state-of-the-art algorithms demonstrates KIRBY's competitiveness on various benchmarks.
\end{abstract}

\section{Introduction}
Out-of-distribution (OOD) detection is important in safety-critical applications where predictions should not only be accurate on average, but also reliable.
Deep neural networks (DNNs) excel at classifying samples drawn from a distribution matching the training set's, but they also tend to inaccurately classify out-of-distribution (OOD) data with high confidence even when the samples deviate significantly from the training distribution \cite{Baseline,SNGP}.
In order to prevent classifiers from producing such predictions, safety-critical applications use OOD detection to alert the user when it is likely that the DNN cannot reliably classify a given input \cite{medicalood_01,autodriving_01,autodriving_02}.

Denote $\scriptX$ to be the set of all possible, including undesirable, input images to a classifier.
An in-distribution (ID) is the distribution from which a training set is sampled.
Its support is described as an ID set $\scriptX_{\text{ID}} \subset \scriptX$, and OOD data $\scriptX_{\text{OOD}}$ is any data that is unlikely to be drawn from this ID.
The task of detecting OOD samples is a binary hypothesis test
\begin{align}
    \Psi\left(x\right) = 
    \begin{cases}
        0 & \text{if } x \in \scriptX_{\text{ID}} \\
        1  & \text{if } x \in \scriptX_{\text{OOD}}
    \end{cases}
\end{align}
where the main difficulty lies in $\scriptX_{\text{OOD}}$ being too large to be represented by any reasonable dataset size.

\begin{figure}[t]
\centering
\includegraphics[width=8.0cm]{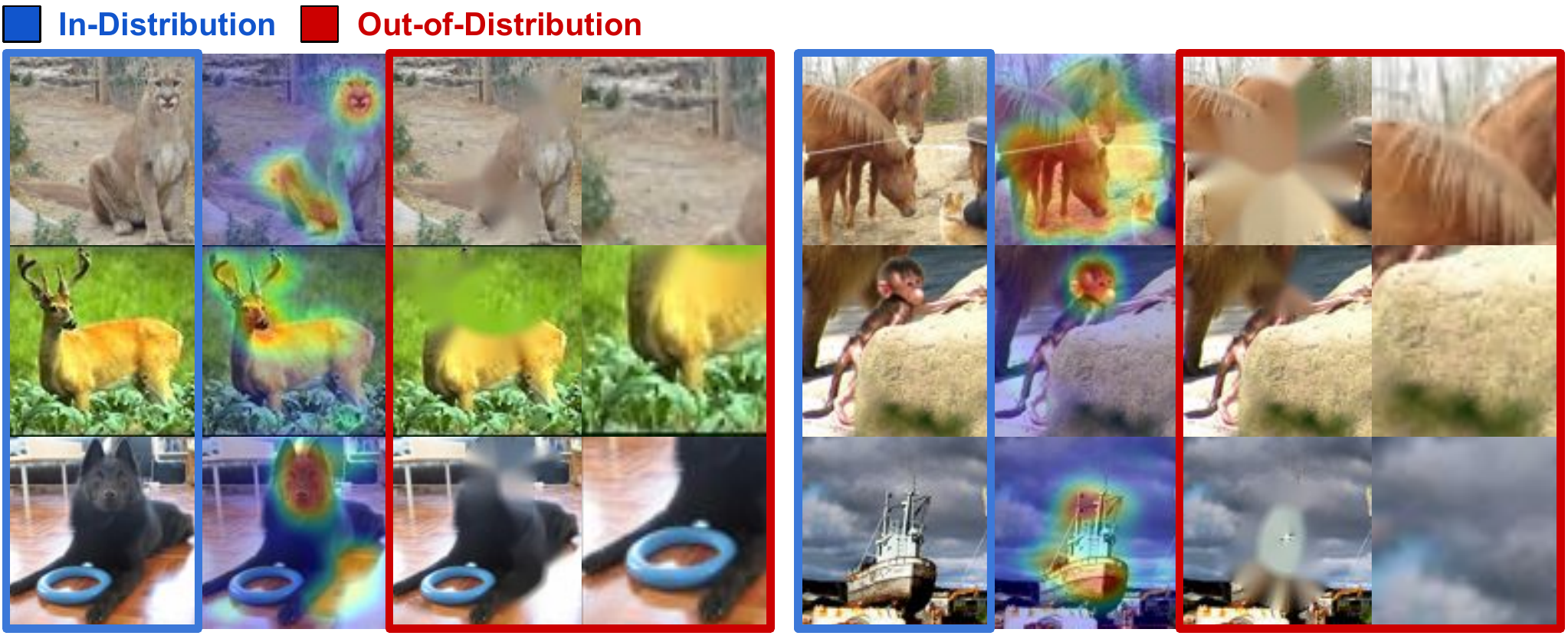}
\caption{Examples of our auxiliary OOD dataset constructed from STL-10 \cite{stl10}.
Each column represents raw images, activation maps, and OOD constructions, respectively.}
\label{fig:stl_result}
\end{figure}

In this paper, we primarily focus on methods to detect OOD samples without having to modify a classifier trained on ID data $\scriptX_{\text{ID}}$.
Having to (re-)train a model specifically for OOD detection on a large training set is computationally demanding, and classification performance may be adversely affected \cite{DUQ}.
Instead, we describe a method to construct an auxiliary dataset of surrogate OOD data from any training (ID) set.

Our work builds upon the observation that OOD samples used to \emph{learn} a decision rule $\Psi$ are most effective when they are semantically similar to ID samples \cite{GAN_outlier}.
First, we establish that OOD samples are perceived by a DNN to be semantically similar to ID samples when they share a common background.
Then we propose a procedure to construct an auxiliary OOD `rejection'' set by replacing class-discriminative features of a training set with marginal background features (see Figure \ref{fig:stl_result}).
A shallow rejection network $\hat{\Psi}$ operating on a pre-trained classifier's latent space can then be trained on the training (ID) and auxiliary OOD datasets to detect OOD data.

Numerical experiments confirm that our procedure indeed generates surrogate OOD data close to ID examples. 
Accordingly, a rejection network trained on this construction outperforms state-of-the-art OOD detection algorithms on most benchmarks.
The benefits of our method are that (1) the OOD dataset is constructed offline while being adaptive to any ID training data; (2) its resultant dataset is ``close'' to ID data which is observed to impact OOD detection performance; (3) a classification network need not be re-trained.
Furthermore, our method does not rely on an ``oracle'' access to a subset of OOD samples.
This contrasts other OOD detection algorithms whose performance degrades when a subset of real OOD data is unavailable \cite{godin,Shafaei2019ALB}.

\begin{figure*}[t]
\centering
\includegraphics[width=16.5cm]{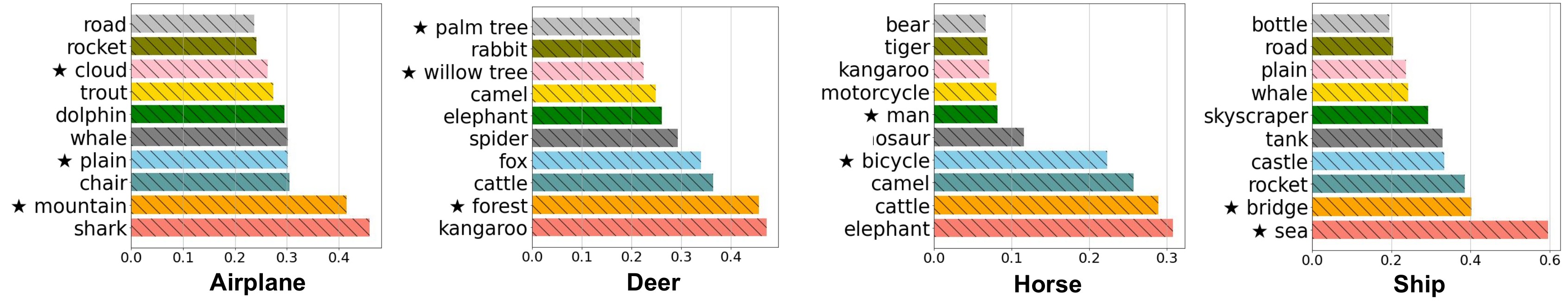}
\caption{Average top-10 confidences of a CIFAR-10 classifier tested on CIFAR-100 data whose classes are listed in rows. For example, a ``sea'' image from CIFAR-100 is classified as ``ship'' with $\approx 0.6$ confidence. Each star ($\star$) indicates an OOD class (CIFAR-100) that has a similar ID background and more results with other ID classes are in Appendix A.
}
\label{fig:research_q}
\end{figure*}

\begin{figure}[t]
\centering
\includegraphics[width=7.5cm]{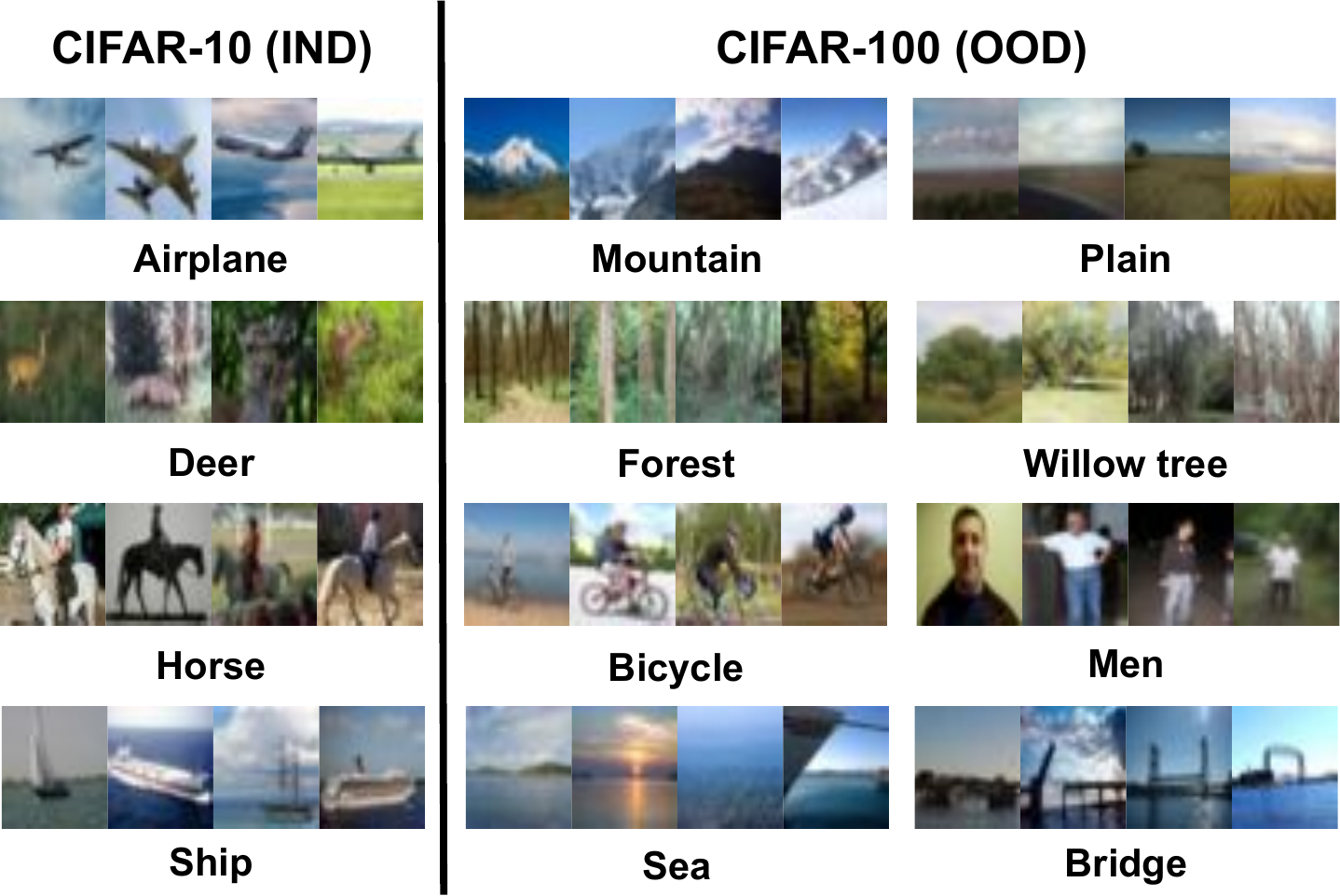}
\caption{OOD samples drawn from right OOD classes are classified as left ID classes with high confidence (MSP) when backgrounds overlap substantially, demonstrating the inadequacy of MSP for OOD detection.
}
\label{fig:background_qualitative}
\end{figure}

\section{Related Work}

Post-hoc algorithms are those that do not require modifying a pre-trained classifier to detect OOD samples.
The simplest post-hoc algorithm thresholds the classifier's maximum softmax probability (MSP) to detect OOD samples \cite{Baseline}.
A substantial improvement was achieved by ODIN \cite{ODIN}, where controlled noise is added to test samples and temperature scaling is applied to better separate predictions on ID and OOD samples.
Treating the latent space of pre-trained models as class-conditional Gaussian variables, \citet{Mahalanobis} uses the Mahalanobis distance with sample mean and covariances as a rejection rule.
Used either as a post-hoc or fine-tuning method, Energy \cite{energy} maximizes the log-sum-exp potential to calibrate classifiers \cite{jem} whose confidence can then be used to detect OOD cases.
ReAct \cite{react} suggests truncating the high activations to address distinctive patterns arising when processing OOD data.
DICE \cite{DICE} is a sparsification technique which ranks weights by contribution, and then uses the most significant weights to reduce noisy signals in OOD data.

Post-hoc methods usually require tuning a parameter on a small subset of OOD data.
Depending on the application, using a subset of true OOD data may be impractical.
Except for MSP and Energy described above, other methods specify parameter(s) that must be tuned on a reserved OOD subset.

\begin{figure*}[t]
\centering
\includegraphics[width=17cm]{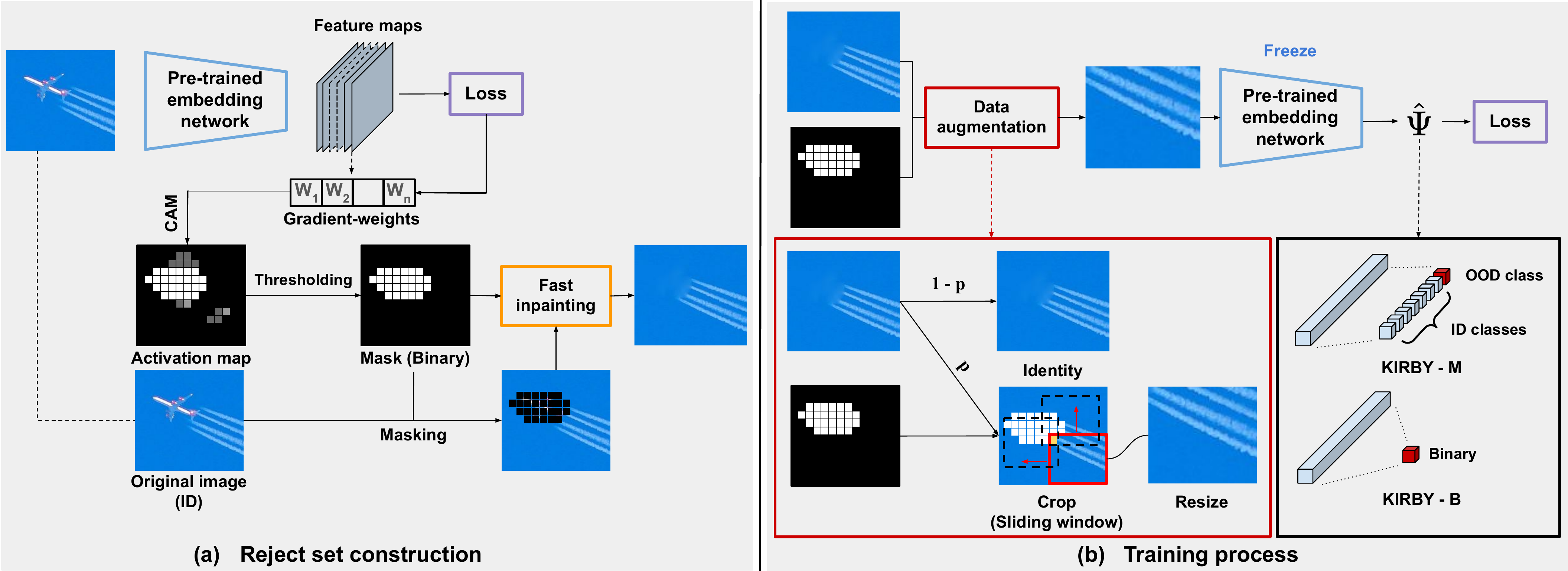}
\caption{
Diagram of KIRBY. (a) Training images are fed into a pre-trained model to locate and erase key ID class-discriminative features. 
A rejection set is then constructed by inpainting the erased images with marginal features.
(b) A rejection network $\hat{\Psi}$ is trained to classify ID and OOD samples, implemented as either KIRBY-M where a multi-class classifier is trained to classify all $K$ classes in the ID set in addition to the OOD class, or KIRBY-B where a binary classifier differentiates ID vs. OOD.
}
\label{fig:aug_ex}
\end{figure*}

Closest to our work are algorithms that train a rejection network on surrogate OOD data.
Outlier Exposure (OE) \cite{OE} uses a static dataset independent of ID samples by distilling the ``80 Million Tiny Images'' dataset \cite{torralba08} and excluding ID samples.
CEDA and ACET \cite{PGD_outlier} both use random noise and pixel shuffling of ID samples, the latter including an adversarial enhancement procedure. 
An analysis in \cite{GAN_outlier} reveals that synthetic examples are most useful as OOD surrogates when they lie near ID examples, and the authors train a classifier with an additional Kullback-Leibler (KL) penalty enforcing uniform predictions on OOD images generated by a generative adversarial network (GAN; \citealp{GAN}).

Excessively distant synthetic examples from ID samples may not help with OOD detection because easy-to-learn outlier features can be discriminated rather trivially.
As described earlier, OOD detection is difficult because the set of OOD samples cannot be covered entirely.
A desirable trait in constructing reject sets is to carefully generate images that do not contain ID classes but are sufficiently close to effectively train a rejection network.
This concept will be revisited later, where the proposed algorithm is shown to satisfy the above trait.

\section{Method}

\subsection{Motivation and Empirical Evidence}

It has been repeatedly observed that surrogate OOD data is most effective when nearby ID data \cite{PGD_outlier,OE,GAN_outlier,ren2019likelihood}.
Conversely, surrogate data far from ID examples hardly affect OOD detection performance \cite{OE}.
This limitation of distant rejection classes is especially evident when OOD detection is evaluated on a set close to ID examples \cite{PGD_outlier}.

\citet{ren2019likelihood,ming2022impact} suggest that neural networks incorrectly classify OOD samples with high confidence when their background overlaps with ID examples.
To complement their observation, we trained a WideResNet \cite{wideresnet} on CIFAR-10 \cite{CIFAR} and plot the average confidence (bars) of predictions on OOD samples (row) that are classified as respective classes in Figure \ref{fig:research_q}.
For example, ``Mountain'' images are on average classified as ``Airplane'' with confidence $\approx 0.4$ whereas $0.1$ is ideal.
Here we observe that softmax confidence scores on OOD classes marked with stars are high when ID classes contain similar background images (see Figure \ref{fig:background_qualitative}).

Based on these observations, we propose a \textbf{K}ey \textbf{I}n-distribution feature \textbf{R}eplacement \textbf{BY} inpainting (KIRBY) procedure to construct surrogate OOD data $\tilde{\scriptX}_{\text{OOD}}$ from ID samples by marginalizing their key representative features with their marginal background features.
A shallow rejection network $\hat{\Psi}$ attached to the pre-trained classifier's layer preceding global average pooling (GAP) is then trained to classify ID samples and the auxiliary OOD dataset.

\subsection{Out-of-Distribution Set Construction}
Let $x \in \scriptX_{\text{ID}}$ be a classifier's input of training samples and $y \in \scriptY = \{1,...,K\}$ be a label.
To generate surrogate OOD samples, we first erase key features from ID training images: 
\begin{equation}
    \tilde{x}_{\text{OOD}} = x \odot \mathbf{M},
\end{equation}
where $\odot$ is element-wise multiplication, and $\mathbf{M} \in \{0,1\}^{W \times H}$ is a binary mask indicating class-specific key regions. 
This mask is computed by thresholding the output $\mathbf{A} \in [0,1]^{W \times H}$ of any class activation map (CAM) algorithm \cite{CAM,GradCAM}:
\begin{equation}
    \mathbf{M}_{ij} = 
    \begin{cases}
        0  & \text{if } \mathbf{A}_{ij} \geq \lambda, \\
        1  & \text{otherwise,}
    \end{cases}
\end{equation}
where $\lambda$ is the threshold that determines how many key features are erased.

Masked regions are then replaced by marginal features using the fast marching inpainting method \cite{inpaint} to produce perceptually plausible OOD images.
The overall pipeline of KIRBY is illustrated in Figure \ref{fig:aug_ex}a.
In the experimental results section, we will explore how choices of CAM and inpainting algorithms affect OOD detection performance.

\begin{table*}[h]
\begin{adjustbox}{width=17.0cm,center}
\centering
    \begin{tabular}{c|c|c|cccccc|cc} 
    \toprule
     &  & WSOL &  \multicolumn{2}{c}{\textbf{CIFAR-10}} & \multicolumn{2}{c}{\textbf{CIFAR-100}} & \multicolumn{2}{c}{\textbf{STL-10}} & \multicolumn{2}{c}{\textbf{Average}}
     \\
    Method & Block \# & Accuracy & FPR $\downarrow$ & AUROC $\uparrow$ & FPR $\downarrow$ & AUROC $\uparrow$ & FPR $\downarrow$ & AUROC $\uparrow$ & FPR $\downarrow$ & AUROC $\uparrow$ \\ 
    \midrule
    \textbf{Smooth-Grad} \cite{SmoothGrad} & Input & - & 15.32 & 96.81 & 33.92 & 91.87 & 71.97 & 79.25 & 40.40 & 89.31 \\ \hline
    \multirow{3}{*}{\textbf{Grad-CAM} \cite{GradCAM}} 
     & 1 & \multirow{3}{*}{53.99} & 17.67 & 96.33 & 37.38 & 90.90 & 52.30 & 87.45 & 35.78 & 91.56 \\
     & 2 &  & 18.36 & 95.96 & 37.91 & 91.09 & 49.37 & 88.34 & 35.21 & 91.79 \\
     & 3 &  & 14.27 & 96.96 & \textbf{33.39} & \textbf{92.41} & 44.92 & 89.10 & 30.86 & 92.82\\ \hline
    \multirow{3}{*}{\textbf{Grad-CAM++} \cite{GradCAMPP}} 
     & 1 &.\multirow{3}{*}{56.42} & 22.98 & 94.79 & 38.92 & 90.72 & 53.84 & 87.10 & 38.58 & 90.87 \\
     & 2 &  & 20.59 & 95.65 & 42.13 & 89.92 & 50.68 & 87.77 & 37.80 & 91.11 \\
     & 3 &  & 14.02 & 97.04 & 34.15 & 92.18 & 44.95 & \textbf{89.47} & 31.04 & \textbf{92.89} \\ \hline
    \multirow{3}{*}{\textbf{Layer-CAM} \cite{LayerCAM}} 
     & 1 & \multirow{3}{*}{57.83} & 20.72 & 95.47 & 40.65 & 90.08 & 53.23 & 87.37 & 38.20 & 90.97 \\
     & 2 &  & 19.72 & 95.75 & 40.34 & 90.34 & 50.34 & 88.02 & 36.80 & 91.37 \\
     & 3 &  & \textbf{14.01} & \textbf{97.03} & 33.64 & 92.31 & \textbf{44.85} & 89.50 & \textbf{30.83} & 92.86 \\ 
     \bottomrule
    \end{tabular}
\end{adjustbox}
\caption{Influence of CAM variants and where features are activated on downstream OOD detection performance for KIRBY-M. Block \# describes the block in WideResNet-40-2 at which CAM is applied. Listed WSOL accuracies are those reported in \cite{LayerCAM} for ILSVRC \cite{ILSVRC}.
}
\label{tab:ablation_cam}
\end{table*}
\subsection{Training a Rejection Network}
A rejection network $\hat{\Psi}$ is trained to classify between ID and (surrogate) OOD samples.
This network can be either a binary classifier (KIRBY-B) or a multi-class classifier (KIRBY-M) that learns an OOD class in addition to the $K$ ID classes.
For KIRBY-M, OOD predictions are obtained by thresholding the reject class at inference time.

When training this network, we augment the surrogate OOD set by randomly choosing to select the patch most distant from the inpainted area, in place of the original surrogate image.
This patch is found with a sliding window that is $1/4$ the size of the original image.
This procedure compensates for artifacts that may be introduced by inpainting, and is illustrated in Figure \ref{fig:aug_ex}.

There are several benefits that come from our method.
Instead of fine-tuning a pre-trained classifier, we attach the rejection network to the penultimate layer of the pre-trained classifier and train the rejection network to leave classification performance unaffected for samples that pass the OOD detection test.
Compared to most post-hoc algorithms, our method does not require a small subset of the OOD set to tune hyper-parameters.
This is critical in many practical applications where the source of OOD samples is unknown and cannot be sampled prior to deployment.

\section{Experiments}\label{sec:Experiments}

\subsection{Setup}\label{sec:setup}
\subsubsection{Datasets}
Following \cite{ODIN,energy}, we test OOD detection performance using CIFAR-10 and CIFAR-100 as ID sets and the following OOD sets:
\begin{itemize}
    \item SVHN \cite{SVHN}. This dataset consists of $32 \times 32$ RGB images associated with digit ($0$--$9$) classes. We use 26,032 test images.
    \item Textures \cite{Textures} was designed as a basis for describable texture attributes, and contains 47 classes spanned across 5,640 images.
    \item LSUN-crop \& LSUN-resize are released by authors from \cite{ODIN}, where the Large-scale Scene UNderstanding dataset (LSUN, \cite{LSUN}) is processed with random crop and downsampling.
    \item Place-365 \cite{Places} test set contains 900 photographs per (scene) class.
    \item iSUN \cite{iSUN} used for OOD detection contains 8,925 images constructed for eye tracking.
\end{itemize}
Additionally we test on STL-10 as ID with a larger resolution of $96 \times 96$ where all OOD examples are re-sized accordingly.
Specifically for both LSUN variants, we follow the procedure in \cite{ODIN} to obtain $96 \times 96$ from the SUN test set instead of re-sizing from their cropped and downsampled versions.

\subsubsection{Evaluation Metrics}
Performance is measured with respect to the following standard criteria.
\begin{itemize}
\item \textbf{FPR} measures the false positive rate (FPR) at the operating point when a negative OOD example is mis-classified as a positive in-distribution sample with true positive rate (TPR) 95\%. Lower is better.
\item \textbf{AUROC} is the area under the receiver operating characteristic curve obtained by varying the operating point. Higher is better.
\end{itemize}

\subsubsection{Training Details} 
A wide variety of modern network architectures were used to compare algorithms: WideResNet-40-2 \cite{wideresnet}, DenseNet-BC (depth $L=100$, growth rate $k=12$) \cite{densenet}, and ResNet-34 \cite{RESNET}.
The rejection network is implemented as two fully-connected layers with its hidden layer's width being $2048$.
Training converges in 5 epochs using SGD-momentum with the initial learning rate of 0.01 and weight decay $5 \times 10^{-4}$.
The rejection network is optimized with cross-entropy loss.
The threshold parameter $\lambda$ is set as $0.3$, and the augmentation that replaces an image with a patch is applied to each sample with probability $0.5$.~\footnote{Code is available at \url{https://github.com/vuno/KIRBY}}

\subsubsection{Baselines}
We compare our method with post-hoc methods: MSP~\cite{Baseline}, ODIN~\cite{ODIN}, Energy~\cite{energy}, Mahalanobis~\cite{Mahalanobis}, ReAct~\cite{react}, and DICE~\cite{DICE} with their hyper-parameters found searching over grids suggested in respective references when necessary; MSP and Energy do not have parameters to tune.
Likelihood-based methods (JEM; \citealp{jem} and LR; \citealp{LR}) are also compared, where the input density $p\left(x\right)$ of a sample $x$ is modeled and used to derive the decision rule $\Psi$.
JEM suffers from training instability, so we follow their experiments and use WideResNet-28-10 without BatchNorm \cite{batch_norm} and additionally report performance using WideResNet-40-2 (without BatchNorm).
Lastly, we compare KIRBY with ACET \cite{PGD_outlier} and GAN \cite{GAN_outlier} which directly learn a rejection rule from surrogate OOD samples as described earlier.
Excluding post-hoc methods, we report averaged AUROC and FPR over five runs.

\begin{figure*}[t]
\centering
\includegraphics[width=15.0cm]{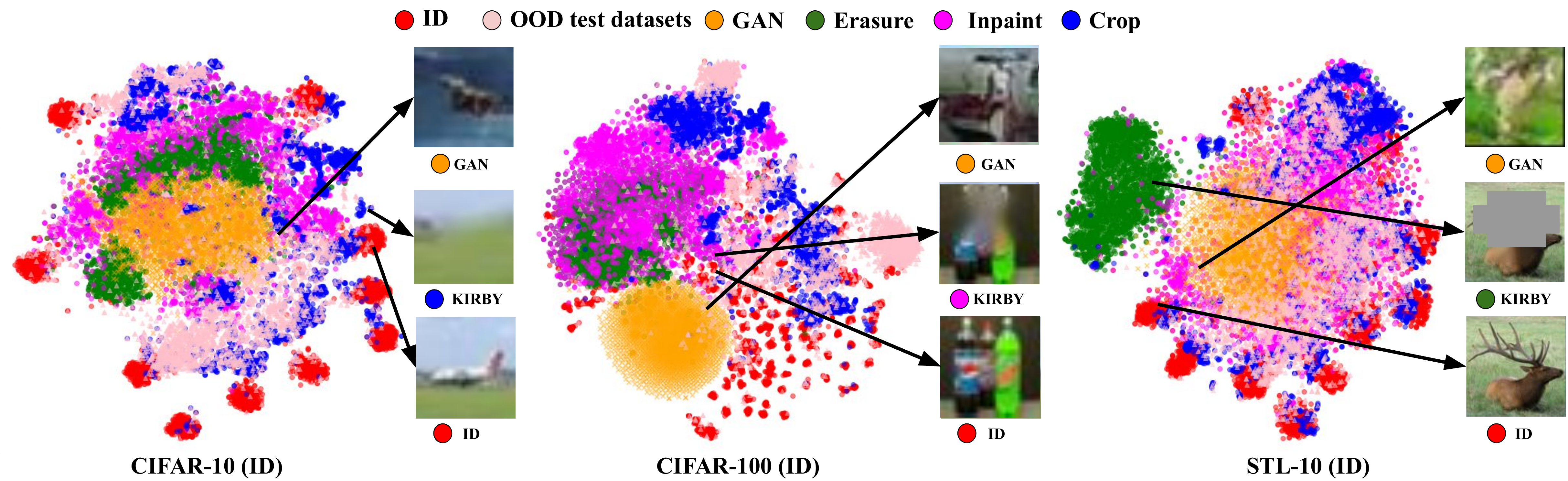}
\caption{Visualization of proximity between auxiliary dataset constructions and ID samples using t-SNE~\cite{tsne}. For each ID sub-figures, all six OOD test sets are used in the above illustration.}
\label{fig:tsne_result}
\end{figure*}

\begin{table}[]
\begin{adjustbox}{width=8.0cm,center}
\begin{tabular}{l|ccc} \toprule
AUROC / FPR & CIFAR-10 & CIFAR-100 & STL-10 \\ \midrule
w/o inpainting & 95.30 / 23.03 & 88.81 / 41.22 & 84.28 / 55.10 \\ 
Mean & 96.59 / 15.82 & 92.07 / 34.05 & 87.34 / 50.58 \\
CS & \textbf{97.40} / \textbf{11.42} & 92.28 / \textbf{29.56} & 84.59 / 51.42 \\
FM & 97.03 / 14.01 & \textbf{92.31} / 33.64 & \textbf{89.50} / \textbf{44.85} \\ \bottomrule
\end{tabular}
\end{adjustbox}
\caption{Effect of different inpainting methods on KIRBY-M's OOD detection performance. We report AUROC and FPR averaged over OOD test sets. }
\label{tab:inpaint}
\end{table}

\begin{table}[]
\begin{adjustbox}{width=8.0cm,center}
\begin{tabular}{l|ccc} \toprule
Latency & CIFAR-10 & CIFAR-100 & STL-10 \\ \midrule
CS & 124.65 $\pm$ 1.32 & 134.65 $\pm$ 1.08 & 1566.87 $\pm$ 10.15 \\
FM & 2.11 $\pm$ 0.51 & 1.84 $\pm$ 0.23 & 8.85 $\pm$ 3.14 \\ \bottomrule
\end{tabular}
\end{adjustbox}
\caption{Computation time (ms / example) comparison between inpainting methods, measured on an Intel(R) Xeon(R) Gold 5220 CPU.}
\label{tab:latency}
\end{table}

\subsection{Design Choices of KIRBY}
We assess in Table \ref{tab:ablation_cam} how the choice of saliency and inpainting methods affect OOD detection performance by experimenting with several CAM and inpainting algorithms.
Additionally, we also show how the choice of layer (residual block) at which features are activated affects performance.
Because saliency maps are often designed for weakly-supervised object localization (WSOL), the WSOL performance is also shown for reference.

We observe that WSOL accuracy is not a good indicator of how the saliency method used in KIRBY would affect OOD detection performance.
Instead, the layer at which CAM is applied is much more important.

\begin{table}[h]
\begin{adjustbox}{width=8.5cm,center}
\begin{tabular}{c|c|c|c|c|c|c} \toprule
 \multirow{2}{*}{CAM} & \multirow{2}{*}{Erasure} & \multirow{2}{*}{Inpaint} & \multirow{2}{*}{Crop} & \textbf{CIFAR-10} & \textbf{CIFAR-100} & \textbf{STL-10} \\ 
 \cline{5-7}
  &  &  &  & AUROC $\uparrow$ & AUROC $\uparrow$ & AUROC $\uparrow$ \\ 
 \midrule
 & \bluecheck &  &  &  95.16 & 88.10 & 83.35  \\
  \bluecheck  & \bluecheck & & & 95.30 & 88.81 & 84.28 \\
  \bluecheck  & \bluecheck & \bluecheck & & 96.33 & 90.07 & 87.76 \\
  \bluecheck  & \bluecheck & \bluecheck & \bluecheck & \textbf{97.03} & \textbf{92.31} & \textbf{89.50} \\ 
  \bottomrule
\end{tabular}
\end{adjustbox}
\caption{Ablation study assessing each component in KIRBY-M. First row assesses the importance of CAM, where instead of identifying key features, it applies a randomly positioned mask covering 25--50\% of input images.
}
\label{tab:ablation}
\end{table}

\begin{table}[h]
\begin{adjustbox}{width=7.8cm,center}
\begin{tabular}{c|ccc} \toprule
Distance $\downarrow$ & CIFAR-10 & CIFAR-100 & STL-10 \\ \midrule
GAN & 86.46 & 203.39 & 408.77 \\ \hline
Erasure (CAM) & 74.48 & 182.30 & 395.24 \\
Erasure+FM & 64.15 & 179.50 & 345.10 \\
Erasure+FM+Crop & 55.00 & 175.88 & 282.16 \\ \bottomrule
\end{tabular}
\end{adjustbox}
\caption{Hausdorff distance between ID and surrogate OOD data. Each component in KIRBY reduces the distance between ID and auxiliary datasets.}
\label{tab:distance}
\end{table}
A surrogate OOD set generated from a shallow layer may not be sufficient to remove entire key features of ID classes since the feature maps from an earlier layer tend to be activated in local features (e.g., edge, and texture). 
Consequently, marginalizing distinctive features at these layers is not as effective as marginalizing higher-level key features. 
In the remaining experiments, we use Layer-CAM at the layer preceding GAP.

Table \ref{tab:inpaint} compares OOD detection performance when using different inpainting: mean pixel value, conditional sampling (CS; \citealp{conditional_sampling}), and fast marching (FM) \cite{inpaint}.
Overall, inpainting methods that fill the masked area by the boundary pixel values around the key feature regions show the best performances.
Unless otherwise specified, we use FM method because FM is 60 - 180$\times$ faster than CS per example (Table \ref{tab:latency}).

\begin{table*}[h!]
\renewcommand\arraystretch{1.0}
\begin{adjustbox}{width=15.0cm,center}
\begin{tabular}{c|c|cccccccccccc} \toprule
\multirow{4}{*}{\textbf{ID}} & \multirow{4}{*}{\textbf{Method}} & \multicolumn{12}{c}{\textbf{OOD Datasets}} \\
 &  & \multicolumn{2}{c}{\textbf{SVHN}} & \multicolumn{2}{c}{\textbf{Textures}} & \multicolumn{2}{c}{\textbf{LSUN-crop}} & \multicolumn{2}{c}{\textbf{LSUN-resize}} & \multicolumn{2}{c}{\textbf{Place-365}} & \multicolumn{2}{c}{\textbf{iSUN}} \\
 &  & FPR & AUROC & FPR & AUROC & FPR & AUROC & FPR & AUROC & FPR & AUROC & FPR & AUROC \\ 
 &  & $\downarrow$ & $\uparrow$ & $\downarrow$ & $\uparrow$ & $\downarrow$ & $\uparrow$ & $\downarrow$ & $\uparrow$ & $\downarrow$ & $\uparrow$ & $\downarrow$ & $\uparrow$ \\ \midrule
\multirow{10}{*}{\textbf{CIFAR-10}} 
 & MSP &  48.43 &  91.91 &  59.11 &  88.51 &  25.52 &  96.48 &  53.39 &  91.07 & 57.04 &  89.52 &  50.11 &  91.18 \\
 & ODIN & 20.10 & 94.70 & 59.11 & 88.51 & 4.37 & 99.04 & 22.50 & 95.13 & 36.63 & 91.78  & 28.29 & 93.97 \\
 & Energy & 35.35 & \textit{91.07} & 52.51 & \textit{85.34} & 4.41 & 99.05 & 28.91 & 93.82 & 34.63 & 91.85 & 31.74 & 92.24 \\
 & ReAct & 36.81 & \textit{90.83} & 51.43 & \textit{87.44} & 5.24 & 98.91 & 31.39 & 93.54 & 35.93 & 90.77 & 37.34 & 92.19 \\
 & Mahalanobis & 6.71 & 98.58 & 17.76 & \textbf{96.53} & 22.06 & 96.47 & 31.05 & 94.98 & \textit{74.05} & \textit{82.11} & 30.68 & 94.67 \\ 
 & DICE & 36.09 & \textit{89.55} & 52.35 & \textit{83.35} & \textbf{1.81} & \textbf{99.59} & 27.74 & 93.87 & 36.65 & 90.59 & 33.22 & 92.43 \\ 
 
 \cline{2-14}
 
 
 & GAN & \textit{86.63} & \textit{77.15} & \textit{84.87} & \textit{72.95} & \textit{88.44} & \textit{71.56} & \textit{76.77} & \textit{80.92} & \textit{75.68} & \textit{80.16} & \textit{80.35} & \textit{78.43} \\
 & ACET  & \textit{56.21} & \textit{91.20} & 51.98 & 89.41 & \textit{50.64} & \textit{91.55} & 49.54 & 91.22 & 55.80 & \textit{88.07} & 48.85 & 91.30 \\ 
 \cline{2-14}
 & KIRBY-M & 6.31 & 98.51 & 21.72 & 94.87 & 2.63 & 99.39 & 13.13 & 97.54 & 26.02 & 94.34  & 14.23 & 97.55 \\ 
 & KIRBY-B & 4.66 & \textbf{98.99} & \textbf{15.84} & 95.86 & 2.05 & 99.53 & \textbf{5.69} & \textbf{98.66} & \textbf{23.05} & \textbf{95.06} & \textbf{4.96} & \textbf{98.85} \\ 
 \hline

\multirow{10}{*}{\textbf{CIFAR-100}} 
 & MSP & 84.35 & 71.37 & 82.65 & 73.54 & 60.33 & 85.58 & 83.27 & 74.11 & 85.17 & 70.46 & 83.24 & 74.95 \\
 & ODIN & 68.12 & 81.34 & 79.53 & 76.68 & 16.98 & 96.95 & 60.59 & 84.96 & 81.74 & 72.57 & 59.47 & 85.56 \\
 & Energy & \textit{85.61} & 73.87 & 79.85 & 76.29 & 23.07 & 95.88 & 80.94 & 77.67 & 82.33 & 72.32  & 72.43 & 77.93 \\
 & ReAct & 78.05 & 87.45 & 68.36 & 83.46 & 24.41 & 95.06 & 80.76 & \textit{72.75} & 78.78 & 75.12 & 81.59 & \textit{73.41} \\
 & Mahalanobis & 28.42 & 94.66 & 40.05 & 90.28 & \textit{76.49} & \textit{73.97} & 20.61 & 96.08 & \textit{85.83} & \textit{66.91} & 25.09 & 94.69 \\  
 & DICE & \textit{86.68} & 74.14 & 76.64 & 76.65 & 11.70 & \textbf{97.84} & 77.78 & 78.84 & 80.60 & 73.34 & 79.11 & 78.89 \\ 
 
 \cline{2-14}
 
 
 & GAN & \textit{85.75} & 76.07 & \textit{91.50} & \textit{66.23} & \textit{92.25} & \textit{63.99} & \textit{90.08} & \textit{64.62} & \textit{91.16} & \textit{62.84} & \textit{91.09} & \textit{62.06} \\
 & ACET & \textit{73.93} & \textit{81.40} & 80.39 & 76.19 & \textit{79.68} & \textit{77.28} & 78.98 & \textit{72.49} & 82.33 & 72.53 & 78.45 & \textit{73.79} \\ \cline{2-14}
 
 & KIRBY-M & 26.71 & 95.21 & 38.31 & 90.91 & \textbf{11.16} & 97.80 & 29.02 & 94.40 & \textbf{70.03} & 80.70 & 26.63 &  94.85 \\ 
 & KIRBY-B & \textbf{14.96} & \textbf{96.26} & \textbf{32.43} & \textbf{91.30} & 12.50 & 97.46 & \textbf{14.58} & \textbf{97.24} & 72.72 & 78.94 & \textbf{13.03} & \textbf{97.48} \\ 
 \hline

\multirow{10}{*}{\textbf{STL-10}} 
 & MSP & 95.98 & 57.56 & 89.23 & 64.53 & 75.59 & 81.64 & 77.07 & 80.91 & 77.84 & 80.15 & 76.57 & 81.60 \\
 & ODIN & 5.20 & 98.78 & 83.00 & 65.41 & 54.79 & 89.17 & 60.27 & 87.52 & 62.79 & 86.38 & 64.89 & 85.70 \\
 & Energy & 89.60 & 64.03 & 87.71 & 64.44 & 54.90 & 89.05 & 59.47 & 87.71 & 60.57 & 86.66 & 63.98 & 85.78 \\
 & ReAct & 90.06 & 66.48 & 86.73 & 71.21 & 56.09 & 88.14 & 60.24 & 86.99 & 61.55 & 86.09 & 63.40 & 87.56 \\
 & Mahalanobis & 5.13 & 98.52 & \textbf{32.25} & \textbf{90.38} & \textit{88.90} & \textit{70.08} & \textit{87.56} & \textit{70.36} & \textit{89.03} & \textit{67.88} & 70.19 & 84.05 \\  
 & DICE & 83.91 & 70.99 & 82.96 & 66.32 & \textbf{44.20} & \textbf{91.40} & 50.68 & 89.82 & 53.72 & 88.53 & 59.94 & 86.30 \\ 
 
 \cline{2-14}
 
 
 & GAN & \textit{97.16} & \textit{53.46} & \textit{95.08} & \textit{56.52} & \textit{91.98} & \textit{60.10} & \textit{90.46} & \textit{62.64} & \textit{89.37} & \textit{63.08} & \textit{86.95} & \textit{62.13} \\
 & ACET  & 95.50 & \textit{54.15} & 89.14 & 65.13 & \textit{79.59} & \textit{77.25} & 76.57 & \textit{79.35} & 76.38 & \textit{79.11} & \textit{90.65} & \textit{63.23} \\ \cline{2-14}
 
 & KIRBY-M & 2.41 & \textbf{99.23} & 65.58 & 79.22 & 53.46 & 89.54 & 50.49 & 89.97 & 53.93 & 88.47 & 43.22 & 90.59 \\ 
 & KIRBY-B & \textbf{1.01} & 98.90 & 76.29 & 69.64 & 47.16 & 90.58 & \textbf{43.59} & \textbf{91.30} & \textbf{44.31} & \textbf{90.31} & \textbf{36.36} & \textbf{92.01} \\ 
 \bottomrule

\end{tabular}
\end{adjustbox}
\caption{Comparison with state-of-the-art methods using WideResNet-40-2.
All experiments are re-run using procedures suggested in respective papers. 
Due to space constraints, we present results for Likelihood-based methods in Appendix C.
}
\label{tab:main_result}
\end{table*}
\subsection{Effect of the Surrogate OOD Set}
An ablation study in Table \ref{tab:ablation} assesses each component of KIRBY by removing or replacing parts as appropriate.
CAM is assessed by replacing it with a random feature mask, i.e., a randomly drawn mask covering 25--50\% of the image is erased.
Other components are incrementally applied, where performance is enhanced consistently with the addition of each step.
Note that procedures up to \textit{Inpaint} comprise auxiliary dataset construction, and \textit{Crop} is a method used to augment training of the rejection network $\hat{\Psi}$.

\begin{figure}[]
\centering
\includegraphics[width=8.2cm]{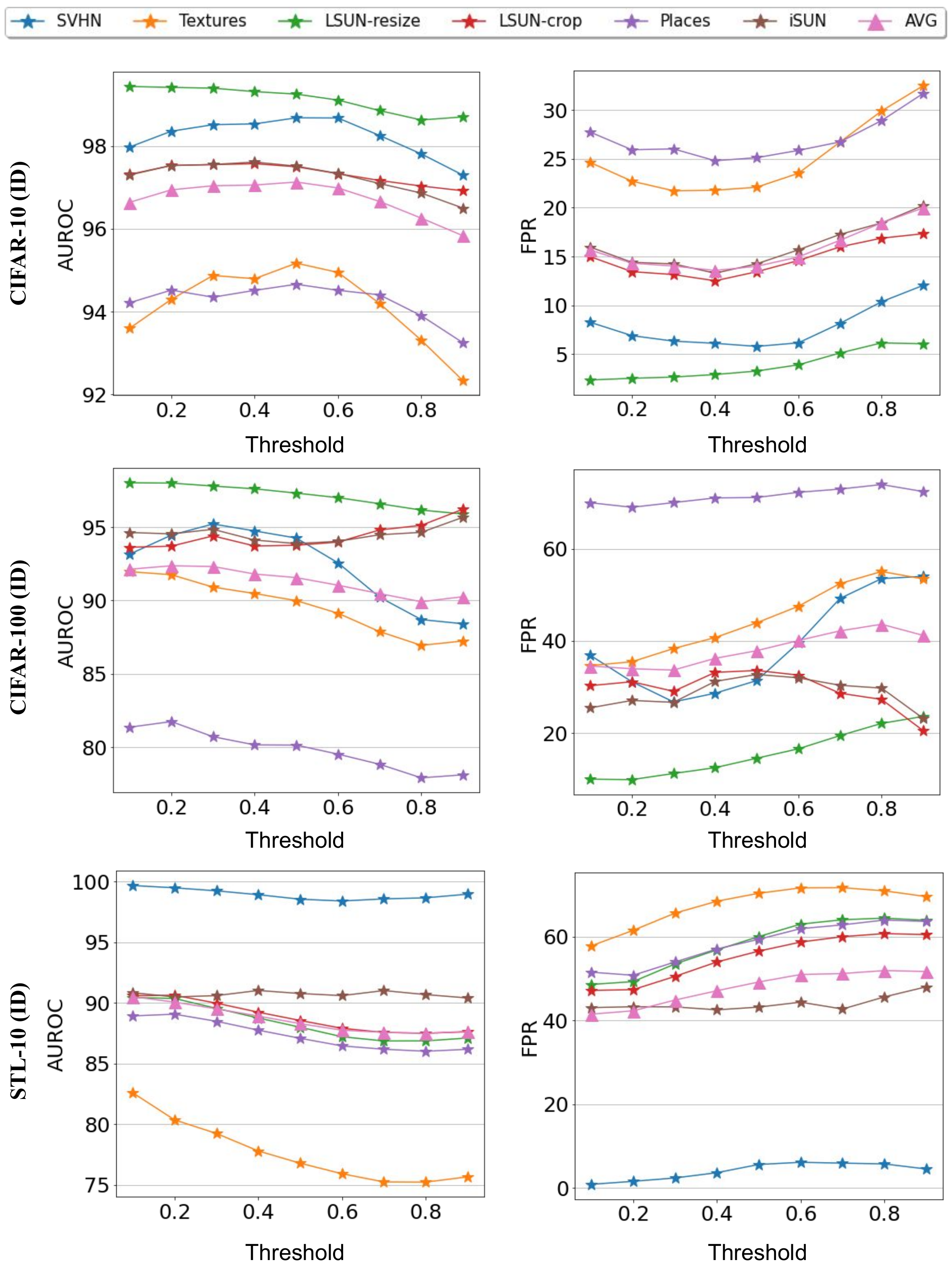}
\caption{KIRBY-M's performances with varying levels of key feature replacement: Increasing $\lambda$ (x-axis) retains more features and affects AUROC and FPR (y-axis).
}
\label{fig:threshold_exp}
\end{figure}

Table \ref{tab:distance} shows that each step indeed contributes to reducing the distance between $\tilde{\scriptX}$ and $\hat{\scriptX}_{\text{ID}}$ as measured by their Hausdorff-Euclidean distance
\begin{equation}
    d_{H} \left(\hat{\scriptX}, \tilde{\scriptX}\right) = \sup_{x \in \scriptX} 
    \lvert \inf_{\hat{x} \in \hat{\scriptX}} d_2 \left(x, \hat{x}\right) - \inf_{\tilde{x} \in \tilde{\scriptX}}d_2 \left(x, \tilde{x}\right) 
    \rvert .
\end{equation}
The Euclidean distance $d_2$ is computed at the classifier's penultimate layer, and a visualization is presented in Figure \ref{fig:tsne_result}.
This is in comparison with the dataset generating using GAN in \cite{GAN_outlier}, where erasure of key features of ID samples is already shown to be closer to the ID set.
Each step enhances OOD detection performance, similar to the earlier observation that surrogate OOD data close to the ID set is beneficial.

The only key parameter that needs to be tuned for KIRBY is the CAM threshold value used to mask key features.
With access to an oracle validation set as assumed by many baselines considered in this work, one could potentially tune this parameter to best suit the downstream OOD set.
Figure \ref{fig:threshold_exp} plots the sensitivity of both evaluation criteria with respect to this threshold parameter, where a larger value implies fewer features being replaced with their background via inpainting.
We identify that both metrics' behavior largely depends on the ID set rather than the OOD set.
This is desirable, as ID examples are necessarily present, whereas OOD samples may not be known a-priori.
Nonetheless, we observe that $\lambda = 0.3$ performs well in general, which we used for the above experiments.
A visualization of inpainting outputs at various thresholds is shown in Appendix B.  
\begin{table*}[h]
\begin{adjustbox}{width=15.0cm,center}
\begin{tabular}{c|ccc|ccc|ccc} \toprule
 & \multicolumn{3}{c}{\textbf{WideResNet}} & \multicolumn{3}{c}{\textbf{ResNet}} & \multicolumn{3}{c}{\textbf{DenseNet}} \\ 
 ID & CIFAR-10 & CIFAR-100 & STL-10 & CIFAR-10 & CIFAR-100 & STL-10 & CIFAR-10 & CIFAR-100 & STL-10 \\ \midrule
MSP & 91.18 & 75.00 & 74.40 & 85.62 & 82.77 & 74.40 & 86.48 & 77.97 & 50.39\\
ODIN & 93.86 & 83.01 & 85.49 & 86.67 & 85.14 & 83.14 & 87.32 & 81.83 & 52.36\\
Energy & 92.24 & 78.99 & 79.61 & 85.63 & 85.23 & 79.19 & 85.75 & 80.97 & 50.42 \\
ReAct & 92.27 & 81.19 & 81.08 & 85.63 & 85.42 & 80.20 & 86.64 & 82.39 & 50.07 \\
Mahalanobis & 93.89 & 86.10 & 80.21 & 95.11 & 82.97 & 66.98 & 90.65 & 84.21 & 74.14 \\
DICE & 91.56 & 79.95 & 82.23 & 90.68 & 72.91 & 83.65 & 93.53 & 86.09 & 84.83  \\
\hline
KIRBY-M & 97.03 & 92.31 & \textbf{89.50} & 96.94 & 88.32 & \textbf{84.74} & \textbf{97.03} & \textbf{93.57} & \textbf{91.25}  \\ 
KIRBY-B & \textbf{97.82} & \textbf{93.11} & 88.79 & \textbf{97.51} & \textbf{90.44} & 82.30 & 96.54 & 92.99 & 88.59  \\ \bottomrule
\end{tabular}
\end{adjustbox}
\caption{OOD detection AUROC using different classifier architectures averaged over the six OOD benchmark datasets.
Additional results are reported in the supplementary material.}
\label{tab:comparison_arc}
\end{table*}
\begin{table*}[]
\renewcommand\arraystretch{1.2}
\begin{adjustbox}{width=16.5cm,center}
\begin{tabular}{c|c|cccccccccccccc} \toprule
\multirow{4}{*}{\textbf{ID}} & \multirow{4}{*}{\textbf{Method}} & \multicolumn{14}{c}{\textbf{OOD Datasets}} \\ 
 &  & 
 \multicolumn{2}{c}{\textbf{SVHN}} & 
 \multicolumn{2}{c}{\textbf{Textures}} & \multicolumn{2}{c}{\textbf{LSUN-crop}} & \multicolumn{2}{c}{\textbf{LSUN-resize}} & \multicolumn{2}{c}{\textbf{Place-365}} & 
 \multicolumn{2}{c}{\textbf{iSUN}} & 
 \multicolumn{2}{c}{\textbf{Average}}\\
 &  & FPR & AUROC & FPR & AUROC & FPR & AUROC & FPR & AUROC & FPR & AUROC & FPR & AUROC & FPR & AUROC \\ 
 &  & $\downarrow$ & $\uparrow$ & $\downarrow$ & $\uparrow$ & $\downarrow$ & $\uparrow$ & $\downarrow$ & $\uparrow$ & $\downarrow$ & $\uparrow$ & $\downarrow$ & $\uparrow$ & $\downarrow$ & $\uparrow$ \\ \midrule
\multirow{3}{*}{\textbf{CIFAR-10}} 
 & OE  & 5.23 & 98.36 & \textbf{12.58} & \textbf{97.77} & \textbf{1.21} & \textbf{99.68} & \textbf{5.57} & \textbf{98.88} & \textbf{19.00} & \textbf{96.58} & 6.39 & 98.79 & \textbf{8.33} & \textbf{98.34} \\ \cline{2-16}
 
 & KIRBY-M & 6.31 & 98.51 & 21.72 & 94.87 & 2.63 & 99.39 & 13.13 & 97.54 & 26.02 & 94.34  & 14.23 & 97.55 & 14.01 & 97.03 \\ 
 & KIRBY-B & \textbf{4.66} & \textbf{98.99} & 15.84 & 95.86 & 2.05 & 99.53 & 5.69 & 98.66 & 23.05 & 95.06  & \textbf{4.96} & \textbf{98.85} & 9.37 & 97.82 \\ \hline

\multirow{3}{*}{\textbf{CIFAR-100}} 
 & OE & 61.79 & 87.66 & 61.96 & 84.39 & 14.55 & 97.38 & 72.24 & 78.53 & \textbf{67.96} & \textbf{81.93} & 73.65 & 77.74 & 58.69 & 84.60 \\ \cline{2-16}
 
 & KIRBY-M & 26.71 & 95.21 & 38.31 & 90.91 & \textbf{11.16} & \textbf{97.80} & 29.02 & 94.40 & 70.03 & 80.70 & 26.63 & 94.85 & 33.64 & 92.31\\ 
 & KIRBY-B & \textbf{14.96} & \textbf{96.26} & \textbf{32.43} & \textbf{91.30} & 12.50 & 97.46 & \textbf{14.58} & \textbf{97.24} & 72.72 & 78.94  & \textbf{13.03} & \textbf{97.48} & \textbf{26.71} & \textbf{93.11} \\ 
 \bottomrule
 
\end{tabular}
\end{adjustbox}
\caption{Comparison between KIRBY and OE when using a WideResNet-40-2 classifier.}

\label{tab:oe_comparison}
\end{table*}


\subsection{Comparison with Baselines}\label{sec:results}
Table \ref{tab:main_result} presents the performance of OOD detection algorithms.
KIRBY outperforms all considered baselines on most in-out distribution pairs, even though it never accesses true OOD data unlike other algorithms \cite{godin,Shafaei2019ALB}. 
On at least one ID-OOD pair (italicized entries), MSP outperforms all algorithms except ODIN and KIRBY despite its simplicity.

Post-hoc algorithms generally perform much better than likelihood methods, especially on the harder STL-10 set.
This is accredited to the dataset's small size (5,000 samples) and its higher resolution where density-based methods are fundamentally limited \cite{nalisnick2018do,nalisnick2020detecting}.
In contrast, KIRBY differs in how it does not rely on learning a high-dimensional representation of the latent space.
Instead, KIRBY erases the low-dimensional structure dictating ID classes, and its performance demonstrates its scalability to higher dimensional structured data.
This is again confirmed in Table \ref{tab:comparison_arc} where algorithms that achieve high performance in the above experiments are further compared when using different classification (backbone) architectures.

Outlier exposure (OE; \citealp{OE}) relies on a much larger-scale auxiliary dataset for its reject class than all auxiliary dataset construction algorithms considered in this work. 
Its construction is static, whereas all others including ours adapt with ID sets and keeps the auxiliary set's size to be the same as ID samples for scalability.
Yet we observe in Table \ref{tab:oe_comparison} that KIRBY outperforms OE by a large margin on CIFAR-100 while it under-performs OE on CIFAR-10 by a small gap.
Recall that the auxiliary dataset drawn by OE is limited to $32 \times 32$ resolution images, and its method does not naturally scale to larger images without additional synthetic constructions (e.g. re-sizing).
Lastly, we do not modify the classification network's parameters, whereas OE requires fine-tuning, and its test accuracy slightly decreases from $94.84 \to 94.80$ and $75.96 \to 75.62$ for CIFAR-10 and CIFAR-100, respectively \cite{OE}.

KIRBY removes key features from ID examples to construct an auxiliary dataset, and it may appear to be limited to datasets whose samples contain non-trivial background features.
To disprove this presumption, we complement our experiments by reporting OOD detection performance in Appendix D on gray-scale pairs with Fashion-MNIST \cite{FMNIST} as ID data, and KIRBY outperforms the compared methods.

\section{Conclusion}
We proposed a simple and intuitive key feature replacement procedure using CAM and inpainting techniques.
Motivated by the observation that OOD surrogates are most effective when semantically similar to ID samples, each step contributes to modifying constructed samples to be more similar to ID samples. 
KIRBY is unique in the sense that it does not learn to generate samples but rather detects and erases low-dimensional features most relevant to classification.
Even though KIRBY constructs an OOD dataset whose size is identical to the training samples for scalability and simplicity, it is shown to outperform OE which distills a much larger set of natural images on CIFAR-100.

\bibliography{aaai23}

\newpage
\onecolumn
\begin{appendix}
\section{Supplementary Material}

\vspace{0.5cm}

\subsection{A. Additional Results for CIFAR-10}
\begin{figure*}[h]
\centering
\includegraphics[width=15cm]{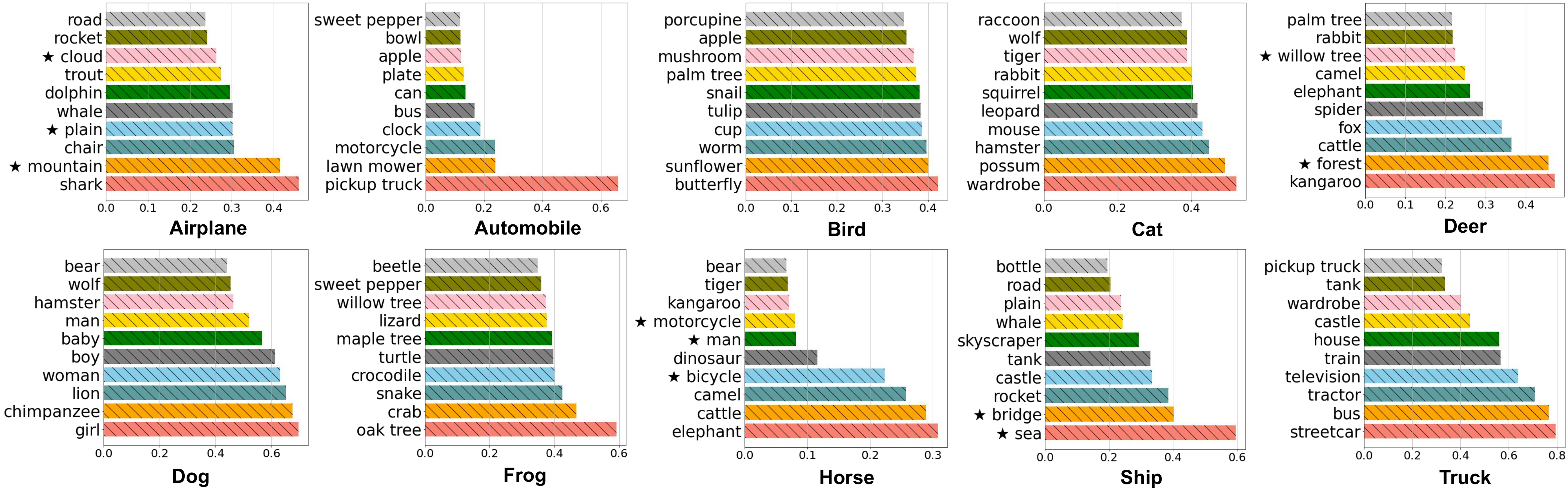}
\caption{Average top-10 confidences of a CIFAR-10 classifier (WideResNet-40-2) tested on CIFAR-100 (OOD).
}
\label{fig:appendix_a}
\end{figure*}

In Figure \ref{fig:appendix_a}, we show additional average softmax probabilities (bars) on OOD samples (row) that are classified as
respective CIFAR-10 classes. 

\subsection{B. Visualization of the Surrogate OOD Samples}
\begin{figure}[h]
\centering
\includegraphics[width=11.8cm]{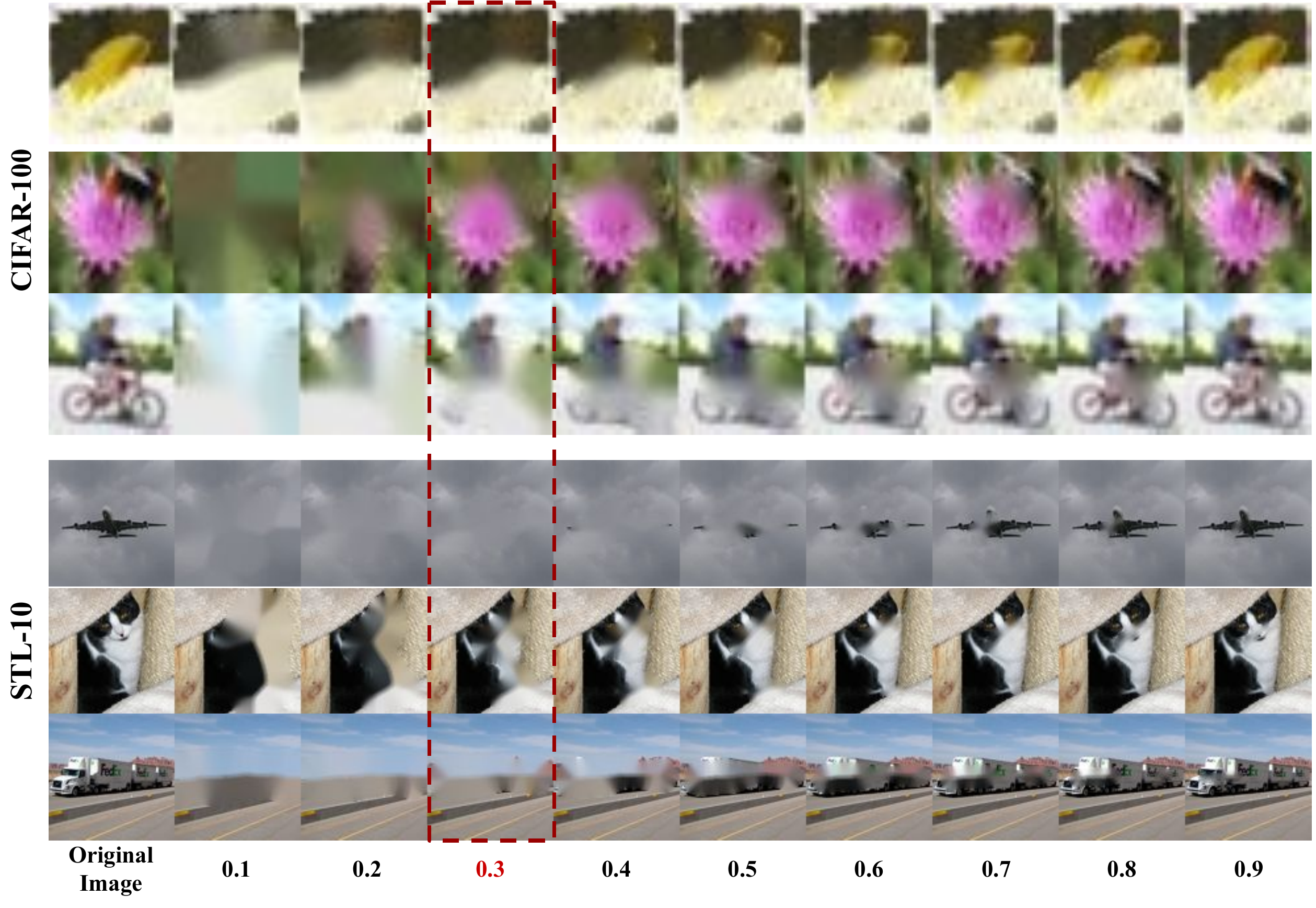}
\caption{Qualitative results for inpainting.}
\label{fig:threshold_qualitative_exp}
\end{figure}
We visualize inpainting results at different CAM thresholds $\lambda$ in Figure \ref{fig:threshold_qualitative_exp}. Key features are incrementally retained with larger thresholds, and we showed that using the threshold of 0.3 generally achieves high OOD detection performance.
Additionally, the surrogate OOD images produced by the threshold of 0.3 are qualitatively reasonable.

\subsection{C. Comparison with Likelihood-based Methods}
We also compare with state-of-the-art likelihood-based methods that incorporated generative modeling \citep{jem,LR} (Table \ref{tab:appendix_b}).
%
\begin{table*}[h!]
\renewcommand\arraystretch{1.0}
\begin{adjustbox}{width=15.0cm,center}
\begin{tabular}{c|c|cccccccccccc} \toprule
\multirow{4}{*}{\textbf{IND}} & \multirow{4}{*}{\textbf{Method}} & \multicolumn{12}{c}{\textbf{OOD Datasets}} \\
 &  & \multicolumn{2}{c}{\textbf{SVHN}} & \multicolumn{2}{c}{\textbf{Textures}} & \multicolumn{2}{c}{\textbf{LSUN-crop}} & \multicolumn{2}{c}{\textbf{LSUN-resize}} & \multicolumn{2}{c}{\textbf{Place-365}} & \multicolumn{2}{c}{\textbf{iSUN}} \\
 &  & FPR & AUROC & FPR & AUROC & FPR & AUROC & FPR & AUROC & FPR & AUROC & FPR & AUROC \\ 
 &  & $\downarrow$ & $\uparrow$ & $\downarrow$ & $\uparrow$ & $\downarrow$ & $\uparrow$ & $\downarrow$ & $\uparrow$ & $\downarrow$ & $\uparrow$ & $\downarrow$ & $\uparrow$ \\ \midrule
\multirow{7}{*}{\textbf{CIFAR-10}} 
 & MSP &  48.43 &  91.91 &  59.11 &  88.51 &  25.52 &  96.48 &  53.39 &  91.07 & 57.04 &  89.52 &  50.11 &  91.18 \\ \cline{2-14}
 
 
 & $\text{JEM}^{\dagger}$-softmax & \textit{80.58} & \textit{85.25} & \textit{69.68} & \textit{87.17} & \textit{75.05} & \textit{84.64} & \textit{62.18} & \textit{88.42} & \textit{68.72} & \textit{85.15} & \textit{64.69} & \textit{87.31} \\
 & JEM-softmax & \textit{82.03} & \textit{83.81} & \textit{68.88} & \textit{87.15} & \textit{81.50} & \textit{80.67} & \textit{72.27} & \textit{85.41} & \textit{73.91} & \textit{83.63} & \textit{72.78} & \textit{85.05} \\
 & JEM-likelihood  & \textit{99.28} & \textit{52.81} & \textit{81.70} & \textit{65.58} & \textit{99.99} & \textit{46.75} & \textit{78.67} & \textit{66.47} & \textit{95.38} & \textit{48.00} & \textit{74.54} & \textit{69.52} \\ 
 & LR  & \textbf{0.00} & 97.25 & 49.99 & 94.22 & \textit{66.66} & \textit{93.32} & 49.99 & 94.39 & \textit{59.99} & 93.38 & \textit{66.66} & 93.01   \\ \cline{2-14}
 
 & KIRBY-M & 6.31 & 98.51 & 21.72 & 94.87 & 2.63 & 99.39 & 13.13 & 97.54 & 26.02 & 94.34  & 14.23 & 97.55 \\ 
 & KIRBY-B & 4.66 & \textbf{98.99} & \textbf{15.84} & \textbf{95.86} & \textbf{2.05} & \textbf{99.53} & \textbf{5.69} & \textbf{98.66} & \textbf{23.05} & \textbf{95.06} & \textbf{4.96} & \textbf{98.85} \\ 
 \hline

\multirow{7}{*}{\textbf{CIFAR-100}} 
 & MSP & 84.35 & 71.37 & 82.65 & 73.54 & 60.33 & 85.58 & 83.27 & 74.11 & 85.17 & 70.46 & 83.24 & 74.95 \\ \cline{2-14}
 
 
 & $\text{JEM}^{\dagger}$-softmax & 73.46 & 81.55 & \textit{85.41} & \textit{70.52} & \textit{77.84} & \textit{75.78} & \textit{86.33} & \textit{70.61} & \textit{85.56} & \textit{67.86} & \textit{88.64} & \textit{66.81} \\
 & JEM-softmax & 78.67 & 78.12 & \textit{89.61} & \textit{62.84} & \textit{90.68} & \textit{60.32} & \textit{86.89} & \textit{67.94} & \textit{89.29} & \textit{63.20} & \textit{88.72} & \textit{64.64} \\
 & JEM-likelihood  & \textit{98.54} & \textit{51.80} & 79.06 & \textit{63.70} & \textit{99.96} & \textit{35.70} & 69.88 & 76.31 & \textit{95.57} & \textit{46.90} & 66.10 & 77.11 \\ 
 & LR & \textit{99.99} & 91.14 & \textit{99.99} & 90.99 & \textit{66.66} & 93.06 & 74.99 & 93.08 & 79.99 & \textbf{92.19} & \textit{83.33} & 92.49  \\ \cline{2-14}
 
 
 & KIRBY-M & 26.71 & 95.21 & 38.31 & 90.91 & \textbf{11.16} & \textbf{97.80} & 29.02 & 94.40 & \textbf{70.03} & 80.70 & 26.63 &  94.85 \\ 
 & KIRBY-B & \textbf{14.96} & \textbf{96.26} & \textbf{32.43} & \textbf{91.30} & 12.50 & 97.46 & \textbf{14.58} & \textbf{97.24} & 72.72 & 78.94 & \textbf{13.03} & \textbf{97.48} \\ 
 \hline

\multirow{7}{*}{\textbf{STL-10}} 
 & MSP & 95.98 & 57.56 & 89.23 & 64.53 & 75.59 & 81.64 & 77.07 & 80.91 & 77.84 & 80.15 & 76.57 & 81.60 \\ \cline{2-14}
 
 
 & $\text{JEM}^{\dagger}$-softmax & 87.32 & 72.96 & 87.84 & 68.37 & \textit{92.38} & \textit{61.85} & \textit{90.82} & \textit{63.80} & \textit{91.67} & \textit{62.73} & \textit{88.27} & \textit{70.19} \\
 & JEM-softmax & 92.67 & 69.83 & \textit{91.90} & \textit{62.73} & \textit{92.39} & \textit{58.62} & \textit{90.73} & \textit{64.56} & \textit{89.19} & \textit{64.94} & \textit{83.22} & \textit{73.38} \\
 & JEM-likelihood  & \textit{99.94} & 57.85 & 82.89 & 67.03 & \textit{95.50} & \textit{50.55} & \textit{95.75} & \textit{57.57} & \textit{96.56} & \textit{55.26} & \textit{98.71} & \textit{62.23} \\ 
 & LR & \textit{99.99} & \textit{27.40} & \textit{99.99} & \textit{54.69} & \textit{99.99} & \textit{66.13} & \textit{99.99} & \textit{72.32} & \textit{99.99} & \textit{69.18} & \textit{99.99} & \textit{64.81} \\ \cline{2-14}
 
 
 & KIRBY-M & 2.41 & \textbf{99.23} & \textbf{65.58} & \textbf{79.22} & 53.46 & 89.54 & 50.49 & 89.97 & 53.93 & 88.47 & 43.22 & 90.59 \\ 
 & KIRBY-B & \textbf{1.01} & 98.90 & 76.29 & 69.64 & \textbf{47.16} & \textbf{90.58} & \textbf{43.59} & \textbf{91.30} & \textbf{44.31} & \textbf{90.31} & \textbf{36.36} & \textbf{92.01} \\ 
 \bottomrule

\end{tabular}
\end{adjustbox}
\caption{Comparison with Likelihood-based methods using WideResNet-40-2. $\text{JEM}^{\dagger}$ is trained with WideResNet-28-10.}
\label{tab:appendix_b}
\end{table*}

\subsection{D. Additional Result}
\subsubsection{D.1. Comparison with representative methods}

\begin{table}[t]
\begin{adjustbox}{width=9.0cm,center}
\begin{tabular}{l|llll} \hline
 & \multicolumn{2}{c}{WideResNet} & \multicolumn{2}{c}{DenseNet} \\ \hline
 & AUROC & FPR & AUROC & FPR \\ \hline
CSI~\cite{CSI} & 92.45 & 35.66 & 85.31 & 47.83 \\
VOS~\cite{VOS} & 94.06 & 24.87 & 95.33 & 22.47 \\ \hline
KIRBY-M & 97.03 & 14.01 & 97.03 & 14.63 \\
KIRBY-B & 97.82 & 9.37 & 96.54 & 15.40 \\ \hline
\end{tabular}
\end{adjustbox}
\caption{OOD detection results with respect to AUROC and FPR, comparing KIRBY with representative methods. All models are trained with CIFAR-10, and we report average AUROC for the six OOD benchmark datasets.}
\label{tab:representative_result}
\end{table}

We provide comparison results with KIRBY and additional related works on feature representation learning (Table \ref{tab:representative_result}).

\subsubsection{D.2. KIRBY Works on Gray-scale Images}
In Table \ref{tab:fmnist_result}, we complement our experiments by reporting OOD detection performance on gray-scale pairs with Fashion-MNIST as ID data. We set MNIST \citep{MNIST}, KMNIST \citep{KMNIST}, and Omniglot \citep{OMNIGLOT} as OOD, respectively.

\begin{table*}[h]
\begin{adjustbox}{width=12.5cm,center}
\centering
\begin{tabular}{c|cccccccc} \toprule
\multirow{4}{*}{\textbf{Method}} & \multicolumn{8}{c}{\textbf{OOD Datasets}} \\ \cline{2-9}
 &  \multicolumn{2}{c}{\textbf{MNIST}} & \multicolumn{2}{c}{\textbf{KMNIST}} & \multicolumn{2}{c}{\textbf{Omniglot}} & \multicolumn{2}{c}{\textbf{Average}} \\
 &  FPR & AUROC & FPR & AUROC & FPR & AUROC & FPR & AUROC  \\ 
 &  $\downarrow$ & $\uparrow$ & $\downarrow$ & $\uparrow$ & $\downarrow$ & $\uparrow$ & $\downarrow$ & $\uparrow$  \\ \midrule
 MSP & 79.00 & 82.53 & 69.45 & 86.61 & 68.82 & 89.36 & 72.43 & 86.17    \\
 ODIN & 48.05 & 91.14 & 41.44 & 91.80 & 43.47 & 94.06 & 44.32 & 92.33   \\
 Energy & 6.31 & 98.85 & 5.89 & 98.80 & 2.96 & 99.32 & 5.06 & 98.99    \\
 ReAct & 4.10 & 99.22 & 3.26 & 99.25 & 2.03 & 99.44 & 3.13 & 99.30   \\
 Mahalanobis & 1.71 & 99.20 & 0.46 & 99.64 & \textbf{0.00} & 99.84 & 0.72 & 99.61   \\ 
 DICE & 0.63 & 99.31 & 0.37 & 99.68 & 0.12 & 99.63 & 0.37 & 99.54   \\ 
 \hline
 
 LR  & 98.99 & 89.26 & 49.99 & 94.31 & 33.33 & 96.00 & 61.11 & 93.19   \\ \hline
 GAN & 53.20 & 89.57 & 52.67 & 89.05 & 97.95 & 51.72 & 67.94 & 76.78    \\
 ACET  & 3.70 & 98.65 & 9.38 & 98.29 & 8.72 & 97.82 & 7.27 & 98.25   \\ \hline
 
 KIRBY-M & \textbf{0.13} & \textbf{99.79} & \textbf{0.22} & \textbf{99.88} & \textbf{0.00} & 99.96 & \textbf{0.12} & \textbf{99.88}    \\
 KIRBY-B  & 5.99 & 98.56 & 1.91 & 99.47 & 0.01 & \textbf{99.99} & 2.63 & 99.34   \\
 \bottomrule
 
\end{tabular}
\end{adjustbox}
\caption{Performance on gray-scale data with WideResNet-40-2. JEM suffers from training instability, hence we exclude JEM in this report.}
\label{tab:fmnist_result}
\end{table*}

\subsection{E. Details of Experiments}

\textbf{Training details.} We use pre-trained WideResNet-40-2~\footnote{\url{https://github.com/facebookresearch/odin}}, ResNet-34~\footnote{\url{https://github.com/pokaxpoka/deep_Mahalanobis_detector}}, and DenseNet-BC~\footnote{\url{https://github.com/facebookresearch/odin}} for CIFAR-10 and CIFAR-100 experiments.
For STL-10, we train classifiers from scratch, and each model is train by minimizing the cross-entropy using SGD with momentum of 0.9.
We use the initial learning rate of 0.1 and the learning rate is dropped by a factor of 10 at 50\% and 75\% of the training progress.
Specifically, we train DenseNet-BC for 300 epochs with batch size of 32.
ResNet-34 and WideResNet-40-2 are trained for 200 epochs with batch size of 128, respectively. 

\noindent \textbf{Software and Hardware.}
We run all experiments with PyTorch \citep{PYTORCH} on a single NVIDIA Tesla V100 GPU.  In particular, we use Fast Marching inpainting method implemented by OpenCV \citep{opencv_library}.

\end{appendix}

\end{document}